\documentclass[journal]{IEEEtran}
 
\usepackage{hyperref}
\usepackage{amsmath,amsfonts,amssymb}
\usepackage{verbatim}
\usepackage[linesnumbered,ruled,vlined]{algorithm2e}
\usepackage{algpseudocode} 
\usepackage{graphicx}
\usepackage{textcomp}
\usepackage{multirow} 
\usepackage{epstopdf}
\usepackage{color}
\usepackage{threeparttable}
\usepackage[table,xcdraw]{xcolor}
\usepackage{subfigure}
\usepackage{changepage}
\usepackage{bm}
\usepackage{diagbox}
\usepackage{booktabs}
\usepackage[switch]{lineno}
\usepackage{pifont}

\newcommand{\mTa}{\mathbf{\Theta}}
\newcommand{\wt}{\mathbf{w}}
\newcommand{\bbC}{\mathbb{C}}
\newcommand{\bht}{\mathbf{h}}
\newcommand{\fd}{\mathrm{d}}
\newcommand{\fr}{\mathrm{r}}
\newcommand{\Gt}{\mathbf{G}}

\begin{document}
\title{AutoOpt: A General Framework for Automatically Designing Metaheuristic Optimization Algorithms with Diverse Structures}

\author{Qi~Zhao,~
        Bai~Yan,~
        Taiwei~Hu,~
        Xianglong~Chen,~
        Jian~Yang,~
        Shi~Cheng,~
	and~Yuhui~Shi,~\IEEEmembership{Fellow,~IEEE}	

\thanks{© 20XX IEEE. Personal use of this material is permitted. Permission from IEEE must be obtained for all other uses, in any current or future media, including reprinting/republishing this material for advertising or promotional purposes, creating new collective works, for resale or redistribution to servers or lists, or reuse of any copyrighted component of this work in other works.}
\thanks{This work is supported by the National Natural Science Foundation of China under Grant No.72401122, Guangdong Basic and Applied Basic Research Foundation under Grants No. 2024A1515012241 and 2021A1515110024, Guangdong University Young Innovative Talents Program Project under Grant No.2024KQNCX050, Fundamental Research Funds for the Central Universities under Grant No. GK202201014, Shenzhen Fundamental Research Program under Grant No. JCYJ20200109141235597, and National Natural Science Foundation of China under Grants No. 61761136008 and 61806119. \textit{(Corresponding author: Yuhui Shi)}}
\thanks{Q. Zhao, J. Yang, and Y. Shi are with the Department of Computer Science and Engineering, Southern University of Science and Technology, Shenzhen 518055, China (email: zhaoq@sustech.edu.cn; yangj33@sustech.edu.cn; shiyh@sustech.edu.cn).}
\thanks{B. Yan is with the School of Computer Science and Technology, Dongguan University of Technology, Dongguan 523808, China (email: yanbai@dgut.edu.cn).}
\thanks{T. Hu is with the Department of Computer Science, Johns Hopkins University, Baltimore, MD 21218, USA (email: thu26@jhu.edu).}
\thanks{X. Chen is with the Dyearn Technology Co. Ltd., Shenzhen, China (email: chenxianglong.cs@outlook.com).}
\thanks{S. Cheng is with the School of Computer Science, Shaanxi Normal University, Xi'an 710119, China (email: cheng@snnu.edu.cn).}}

\maketitle
\begin{abstract}
Metaheuristics are widely recognized gradient-free solvers to hard problems that do not meet the rigorous mathematical assumptions of conventional solvers. The automated design of metaheuristic algorithms provides an attractive path to relieve manual design effort and gain enhanced performance beyond human-made algorithms. However, the specific algorithm prototype and linear algorithm representation in the current automated design pipeline restrict the design within a fixed algorithm structure, which hinders discovering novelties and diversity across the metaheuristic family. To address this challenge, this paper proposes a general framework, AutoOpt, for automatically designing metaheuristic algorithms with diverse structures. AutoOpt contains three innovations: (i) A general algorithm prototype dedicated to covering the metaheuristic family as widely as possible. It promotes high-quality automated design on different problems by fully discovering potentials and novelties across the family. (ii) A directed acyclic graph algorithm representation to fit the proposed prototype. Its flexibility and evolvability enable discovering various algorithm structures in a single run of design, thus boosting the possibility of finding high-performance algorithms. (iii) A graph representation embedding method offering an alternative compact form of the graph to be manipulated, which ensures AutoOpt's generality. Experiments on numeral functions and real applications validate AutoOpt's efficiency and practicability. 
\end{abstract}


\section{Introduction}
The rapid development of modern science and engineering induces hard optimization problems with proprieties of discretization, multi-modality, large-scale, black-box, or multi-objectivity. These problems have to be relaxed or approximated to meet the rigorous mathematical assumptions of conventional optimization solvers. The relaxation and approximation are not straightforward and not always available. Alternatively, the problems can be directly solved by metaheuristics, a class of stochastic search algorithms integrating gradient-free local improvement with high-level strategies of escaping from local optima \cite{glover2006handbook}. Metaheuristics have been widely recognized due to their progressive theoretical foundations \cite{ollivier2017information,fornasier2021consensus} and remarkable performance in various applications, e.g., optimizing representations \cite{schmidt2009distilling,weiel2021dynamic}, model structures \cite{stanley2019designing}, and control decisions \cite{birattari2020disentangling}.

Metaheuristics cover a rich family of algorithms ranging from neighborhood search-based ones to evolutionary and swarm algorithms \cite{glover2006handbook}, \cite{deng2023snow,klein2018cheetah,zhong2022beluga,klein2015modified}. Most of the algorithms further have tunable parameters. Such rich choices of algorithms and parameters incur a high degree of freedom in tailoring an algorithm to a specific problem. Manually trial-and-error over the high degree of freedom could be laborious, time-costly, and prone to overfit. This motivates automating the algorithm design process with little manual effort. By leveraging computing power to replace human experts to conceive, build up, and verify the design choices, the automated design could provide researchers and practitioners quicker and easier access to eligible algorithms for solving their problems \cite{stutzle2019automated,2020The,zhao2023survey}. Furthermore, by fully exploring potential design choices and discovering novelties with computing power, the automated design could go beyond human experience and gain enhanced performance regarding human problem-solving \cite{fawzi2022discovering,chen2023symbolic}. 

The automated design of metaheuristic algorithms relies on three key modules: \ding{182} a design space regulated by an algorithm prototype and algorithm representation, \ding{183} a principled method to produce algorithms by exploring the design space, and \ding{184} a strategy to computationally efficiently evaluate or estimate the performance of the produced algorithms. Many methods (e.g., SMAC's Bayesian optimization \cite{lindauer2022smac3}, irace's estimation of distribution (EDA) \cite{lopez2016irace}, model-free local \cite{hutter2009paramils} and global \cite{koza1994genetic} search, analytic hierarchy process \cite{song2023brick}, autoregressive learning \cite{zhao2024automated}) and strategies (e.g., racing \cite{lopez2016irace}, intensification \cite{hutter2009paramils}, capping \cite{hutter2009paramils,de2022capping}, surrogate model \cite{lindauer2022smac3}) have been devoted to modules \ding{183} and \ding{184}, respectively, and got strong track records \cite{zhao2023survey}. However, the development of module \ding{182}'s algorithm prototype and representation lags well behind, which poses the automated design great challenges in discovering novelties and diversity:
\begin{itemize}
    \item Algorithm prototype regulates a coherent template to form an algorithm, which avoids infeasible algorithms incurred by arbitrarily instantiating from the design space. Current studies use specific prototypes to bias the design toward a certain segment of the metaheuristic family, e.g., using genetic algorithm (GA) \cite{holland1973genetic} prototype to regulate designing GA variants \cite{bezerra2015automatic,bezerra2020automatically,yi2022automated}. This hinders discovering novelties beyond the chosen segment and may lead to sub-optimal problem-solving, since the best type of algorithms is often unknown for real problems. 
    \item Algorithm representation transforms the algorithm expressed with particular languages or implementations into the one that computer programs of modules \ding{183} and \ding{184} can recognize. The common practice is vector representation, in which categorical identifiers represent algorithmic operators and numerals represent inner parameter values. Operators' execution order is normally predefined, and the length of the vector is fixed for ease of manipulation \cite{bezerra2015automatic,bezerra2020automatically,villalon2021pso,yi2022automated}. This indicates a fixed linear ordering of operators with little potential to discover diversified algorithm structures.
\end{itemize}

To address the challenges, in this paper, we propose a general framework, AutoOpt \footnote{Source code available at \url{https://github.com/auto4opt/AutoOpt}.}, for automatically designing metaheuristic optimization algorithms with diverse structures. AutoOpt's main contributions are:
\begin{enumerate}
    \item \textit{General algorithm prototype}. We propose a general algorithm prototype dedicated to covering the metaheuristic family as widely as possible. Various types of algorithms can be instantiated through the prototype, e.g., algorithms with unfolded tandem operators, algorithms with inner loops of local search, and algorithms with multiple pathways. Such a general prototype promotes high-quality automated design on different problems by fully discovering potentials and novelties across the metaheuristic family. 
    \item \textit{Graph algorithm representation}. We develop a directed acyclic graph algorithm representation to fit the proposed prototype. The graph is flexible to represent various algorithm structures instantiated from the prototype. Furthermore, it is straightforward to be manipulated and evolved during algorithm design. The flexibility and evolvability enable discovering diversified algorithms in a single run of design, thus boosting the possibility of finding high-performance algorithms. 
    \item \textit{Graph representation embedding}. In design scenarios with massive candidate design choices, the graph's adjacent matrix is large-scale and sparse; while the adjacent list is variable-length considering different algorithm structures. The large-scale, sparse, and variable-length nature challenges the direct manipulation of the graph. To relieve the issue, we employ graph auto-encoder to learn a compact embedding for the graph. The embedding offers an alternative form to be manipulated, ensuring AutoOpt's generality.  
\end{enumerate}
We investigate AutoOpt's performance on both numerical functions and real applications. Results confirm its efficiency and practicality.

The remainder of the paper contains: preliminaries of the automated design of metaheuristic algorithms in Section \ref{sec_background}, the proposed AutoOpt framework in Section \ref{sec_autoopt}, experiments and applications in Section \ref{sec_experiment}, and conclusions in Section \ref{sec_conclusion}.  

\section{Preliminaries}\label{sec_background}
\subsection{Automated Design of Metaheuristic Algorithms}
Without loss of generality, the design of metaheuristic algorithms can be formulated as
\begin{equation}\label{eq_obj}
    \mathop{\arg\max}\limits_{A=\langle A_{ops},A_{parms}\rangle \in\mathcal{D}} \ \mathbb{E}_{\mathcal{I}}\big [\mathbb{E}_{\mathcal{P}}[P(A|i)] \big], \ i\in\mathcal{I}, P\in\mathcal{P},
\end{equation}
where the designed algorithm $A$ is a tuple $\langle A_{ops},A_{parms}\rangle$, in which $A_{ops}$ and $A_{parms}$ are the operators and parameters that constitute $A$, respectively; $\mathcal{D}$ is the design space, from where $A$ can be instantiated; $i$ is an instance from the target problem domain $\mathcal{I}$; $P:\mathcal{D}\times\mathcal{I}\to\mathbb{R}$ is a metric that scores the performance of $A$ by a run of $A$ on $i$. Because metaheuristic algorithms conduct stochastic search, we need to estimate the expected performance over $\mathcal{P}$, i.e., multiple runs of $A$ result in multiple $P\in\mathcal{P}$. 

The distribution of a real problem's instances is often unknown. The common practice to handle this is targeting a finite set of instances from $\mathcal{I}$. Consequently, Eq. (\ref{eq_obj}) is reformulated as
\begin{equation}\label{eq_obj2}
    \begin{aligned}
    \mathop{\arg\max}\limits_{A=\langle A_{ops},A_{parms}\rangle \in\mathcal{D}} \ & \mathbb{E}_{I_{t}} \big [\mathbb{E}_{\mathcal{P}}[P(A|i)] \big ],\\
    \ i\in I_{t}\subseteq\mathcal{I}, \forall t \in \{1,& 2,\cdots,T\}, \ P\in\mathcal{P},
    \end{aligned}
\end{equation}
where $I_{t}$ is the finite set of problem instances that are targeted at time (i.e., iteration\footnote{Since Equation \ref{eq_obj2} is black-box, it is often solved in an iterative manner.}) $t$ of the design process. The instances can either be fixed (i.e., $I_{1}=I_{2}=\cdots=I_{T}$) or dynamically changed during the design process. The output of solving Eq. (\ref{eq_obj2}) is an algorithm (or algorithms) with the best performance on the instances. To avoid overfitting, the design process should be followed by a test on the designed algorithms' generalization to instances from $\mathcal{I}\backslash \{I_{1},I_{2},\cdots,I_{T}\}$.

The automated design could be computationally intensive due to the large amount of performance evaluations required for the algorithms produced during the design process. Thus, the automated design is suitable to scenarios where one can afford a priori computational resource (for algorithm design) to subsequently solve many problem instances drawn from the target domain. With the rapid advancement of modern computational resources, automated design has become increasingly appealing, offering an efficient alternative to the laborious trial-and-error of human designers, especially considering that human experts may not always be available in practice. 

\subsection{Pipeline of Automated Design}
\begin{figure*}[t] 
	\centering
	\includegraphics[width=0.7\linewidth]{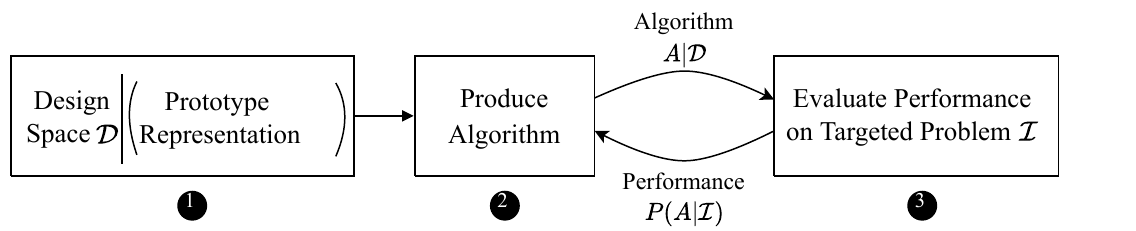}
	\caption{Pipeline for automatically designing metaheuristic algorithms.}
	\label{pipeline}
\end{figure*}

Automated design of metaheuristic algorithms generally follows the pipeline in Figure \ref{pipeline}. The pipeline contains three modules. In module \ding{182}, design space $\mathcal{D}$ provides elementary components for constructing algorithms. There are two types of design space. The first type is with computational primitives, e.g., $+, -, *, /, swap, delete$. The primitives can be involved in designing symbolic formulations of algorithmic operators \cite{richter2018automated,richter2019evolving}. The designed operators can then be composed as a complete algorithm. The second type is with existing algorithmic operators, e.g., 2-opt \cite{lin1965computer}, Gaussian mutation \cite{back1993overview}, and Tabu list \cite{feo1995greedy}. A composition of the operators can be designed to form an algorithm \cite{bezerra2015automatic,bezerra2020automatically,villalon2021pso}. The second type of design space is more popular in the literature, because the design space with operators is more compact than that with computational primitives, and it allows the design inheriting from prior knowledge and experimentation. The compact and knowledge-induced space makes the design easier and available with fewer computing resources than designing from computational primitives \cite{zhao2023survey}. 

Algorithm prototype $\mathcal{C}$ and algorithm representation $\mathcal{R}$ regulate what algorithms can be found from the design space. Arbitrarily instantiating algorithms over the design space may result in infeasible algorithms. To avoid this, the prototype $\mathcal{C}$ defines a coherent template to form an algorithm. For example, simulated annealing (SA) \cite{kirkpatrick1983optimization}, particle swarm optimization (PSO) \cite{shi1998modified}, and GA \cite{holland1973genetic} prototypes were used to regulate designing new variants of SA \cite{franzin2019revisiting}, PSO \cite{villalon2021pso}, and GA \cite{bezerra2015automatic,bezerra2020automatically,yi2022automated}, respectively. The Push language \cite{spector2001autoconstructive} and Backus-Naur form grammar \cite{ryan1998grammatical} with predefined production rules also served as a prototype to map the algorithm representation to an executable algorithm \cite{kamrath2020automated,lones2021evolving,miranda2020novel,alfaro2020automatic}. The major concern is that the specific prototypes limit designing a certain type of algorithms rather than discovering novelties across the metaheuristic family. This may lead to sub-optimal problem-solving, since the best type of algorithms is often unknown for real problems.

The binary tree \cite{koza1994genetic} and vector representations are dominant for the design space with computational primitives and that with operators, respectively. In the binary tree, terminals represent input variables, parameters, or constants; non-terminals represent computational primitives that determine the operations on the terminals. The binary tree is often employed to represent a single algorithmic operator rather than a complete algorithm, because representing a complete algorithm incurs a bloated tree that is non-trivial to be managed \cite{woodward2009evolution}. In vector representation, categorical identifiers represent algorithmic operators; numerals represent inner parameter values. Operators' execution order is normally predefined by the algorithm prototype \cite{bezerra2015automatic,bezerra2020automatically,villalon2021pso,yi2022automated} or associated with the vector \cite{oltean2005evolving}; the length of the vector is fixed for ease of manipulation \cite{bezerra2015automatic,bezerra2020automatically,villalon2021pso,yi2022automated}. Directed graph representation, in which vertices represent operators and directed edges determine the ordering of operators, is more expressive. Cartesian genetic programming (GP) \cite{miller2011cartesian}'s acyclic graph was employed to order operators in a variable-length feed-forward pathway \cite{kantschik1999meta,shirakawa2009evolution,ryser2016iterative}. It, however, still realizes linear serial structures. The directed cyclic graph of \cite{tisdale2021directing} can represent multiple nonlinear pathways by allowing each vertex to be connected from multiple previous vertex. It dedicates to express various $(\mu / p + \lambda)$ and $(\mu / p,\lambda)$ evolutionary algorithms (EAs) but lacks generalization to describe other metaheuristics, e.g., ones with inner loops and ones with operator's self-recursion.

Module \ding{183} produces algorithms over the design space. Various model-based and model-free methods have been developed to produce algorithms. Representative model-based methods are SMAC \cite{lindauer2022smac3} and irace \cite{lopez2016irace}. The former employs Bayesian optimization to infer the algorithm with the best performance on the targeted problem. The latter utilizes estimation of distribution \cite{baluja1995removing} to iteratively sample algorithms from the distribution towards the best algorithms. Model-free methods include ParamILS's \cite{hutter2009paramils} iterative local search (ILS) \cite{lourencco2003iterated}, evolutionary search, and reinforcement learning. ParamILS focuses on local refinement of the current algorithm, given the evidence that performance enhancement is often obtained by adjusting a few hyper-parameters of the algorithm \cite{pushak2022automl}. Evolutionary search, especially GP \cite{koza1994genetic}, is the dominant method of producing algorithms over the design space with computational primitives, because of GP's tree-based search ability. Deep reinforcement learning methods, e.g., deep Q-network \cite{mnih2015human} and proximal policy optimization \cite{schulman2017proximal}, were employed to learn policies of how to produce algorithms given the targeted problem instances \cite{yi2022automated,schuchardt2019learning}.

Module \ding{184} evaluates the produced algorithms' performance, e.g., solution quality, running time, and anytime performance \cite{ye2022automated}, on the target problem. The performance prompts the design toward finding desired algorithms. Thoroughly evaluating algorithms on all targeted problem instances may incur computational overload. Thus, various strategies have been proposed to reduce the number of evaluations (e.g., racing \cite{lopez2016irace} and intensification \cite{hutter2009paramils}) or estimate the performance without a full evaluation (e.g., capping \cite{hutter2009paramils,de2022capping} and low-complexity surrogate models \cite{lindauer2022smac3}). 

\section{AutoOpt}\label{sec_autoopt}
The proposed AutoOpt framework realizes automatically designing metaheuristic algorithms with various structures by (i) a general algorithm prototype, (ii) a directed acyclic graph algorithm representation, and (iii) the graph representation embedding. We first present the three key elements in subsections \ref{sec_prototype}, \ref{sec_representation}, and \ref{sec_embedding}, respectively. Then, we illustrate the overall framework in subsection \ref{sec_framework}. 

\subsection{General Algorithm Prototype} \label{sec_prototype}                                                
\subsubsection{Motivation}
Algorithm prototypes in the literature normally abstract a certain type of metaheuristics, e.g., SA \cite{franzin2019revisiting}, PSO \cite{villalon2021pso}, or GA \cite{bezerra2015automatic,bezerra2020automatically,yi2022automated}. From the problem-solving point of view, they block potentially promising algorithms beyond the chosen type, subsequently may result in sub-optimal algorithms. This motivates us to propose a general prototype that covers the metaheuristic family as widely as possible, which enables a high-quality algorithm design by fully discovering potentials and novelties across the family. 

\subsubsection{Proposed Prototype}
The proposed prototype is shown in Algorithm \ref{alg_prototype}. It begins with the initial solution(s) $S$. $S$ contains decision variables, objective values, and conditional variables induced by the instantiated algorithm, e.g., constraint violations in cases with constraint handling, velocities in cases with PSO's particle fly operator \cite{shi1998modified}, and indexes of associated reference vectors in cases with decomposition-based selection operator for multi-objective problems \cite{zhang2007moea}. In line 1, it checks whether there are \texttt{archive} operators, e.g., archiving elite solutions as done in elite EAs, archiving visited solutions to form a tabu list \cite{feo1995greedy}, and archiving diversified solutions as done in the quality-diversity algorithm \cite{pugh2016quality}, etc. If so, the archive set $A$ is initialized in lines 2 and 3. 
\begin{algorithm}[t]
\caption{Proposed general algorithm prototype} 
\label{alg_prototype}
\KwIn{initial solution(s) $S$}
\If{$\bigcup_{i=1}^{p}\texttt{archive}_{i}\neq\varnothing$}{
    \For{$i=1$ to $p$}{            
    $A_i\leftarrow\varnothing$           
    }        
}        
\While{algorithm not terminate}{
    \For{$j=1$ to $q$}{
        \While{$\texttt{condition}_j$ not met}{
            $S^{'}\gets\texttt{choose}(S)$ \\                
            $S_{new}\gets\texttt{search}_{j}(S^{'}, A)$ \\               
            $S\gets\texttt{update}(S,S_{new})$ \\                
            \If{$\bigcup_{i=1}^{p}\texttt{archive}_{i}\neq\varnothing$}{
                \For{$i=1$ to $p$}{
                    $A_{i}\leftarrow\texttt{archive}_{i}(A_{i},S)$
                }            
            }          
        }            
    }       
}
\KwOut{best solution(s) from $S$ or $A$}
\end{algorithm}

After that, the prototype goes into the main iteration. The \textbf{for} loop in lines 5-12 performs the $q$ search pathways of the instantiated algorithm. The \textbf{for} loop started at line 5 and the $\texttt{condition}_{j}$ in line 6 control instantiating various algorithm structures. For example, a usual unfolded metaheuristic algorithm with a single search pathway can be instantiated by setting $q=1$ and $\texttt{condition}_{1}$ as the algorithm termination condition. A variable neighborhood search with $q$ search operators $\texttt{search}_{j}$, $j=1,2,\cdots,q$, can be instantiated by executing the \textbf{for} loop serially, with  $\texttt{condition}_{j}$ determining switching the $j$th search to $j+1$th. An iterative local search with a local improvement $\texttt{search}_{1}$ and global perturbation $\texttt{search}_{2}$ can be instantiated by executing the \textbf{for} loop serially, with $\texttt{condition}_{1}=``reach \ a \ local \ optima"$ and $\texttt{condition}_{2}=``consume \ |S_{new}| \ function \ evaluations"$, where $|S_{new}|$ counts the number of solutions in $S_{new}$. A memetic algorithm with a local refinement meme operator and a global search operator can be instantiated in the same way. An ensemble algorithm with $q$ parallel search pathways can be instantiated by executing the \textbf{for} loop in parallel, with the first execution of the $\texttt{choose}$ operator dividing $S$ into $q$ subpopulations $S^{'}$.

A common search pathway consists of a \texttt{choose}, \texttt{search}, and \texttt{update} operators, as shown in lines 7-9. The \texttt{choose} operator determines the starting point(s) $S^{'}$ to search from. $S^{'}$ can be identical to $S$ as done in most individual-based search and swarm algorithms, or a segmentation or augmentation of $S$ as done by EAs' mating selections. In line 8, $\texttt{search}_{j}$ searches from $S^{'}$ to construct new complete solution(s) $S_{new}$. $\texttt{search}_{j}$ can be various metaheuristic search operators. The archive set $A$ may be involved, e.g., archiving previously visited solutions to restrict search in unexplored regions. In line 9, the \texttt{update} operator updates $S$ by, e.g. the usual greedy fashion as done in individual-based metaheuristics and elitism EAs, the metropolis fashion as done in SA, a pivoting rule \cite{yannakakis2003local} in cases with multiple $S_{new}$ from scanning the neighborhood of $S$, the pair-wise fashion in cases with population-based search, or always accept $S_{new}$ as done in some swarm algorithms, etc. In lines 10-12, the archive set $A$ is updated by the \texttt{archive} operator. 

\subsubsection{Remarks} 
For the \texttt{choose} operator, some metaheuristics do not have an explicit \texttt{choose} but directly use $S$ as the starting point(s) for search. In that case, \texttt{choose} is the operator identically copying $S$ to $S^{'}$. For the \texttt{search} operator, some search behaviors may not independently act as the \texttt{search}, e.g., an independent $n$-point crossover may not be sufficient for problems excluded from the permutation ones, because it does not introduce novelties beyond what have been involved in current solutions. In that case, several search behaviors collaboratively form the \texttt{search} operator, e.g., a crossover followed by a mutation. Besides, starting from $S^{'}$, some search behaviors may perform multiple rounds iteratively or recursively to generate $S_{new}$. Such iteration and recursion are encapsulated within the \texttt{search} operator; parameters for controlling such iteration or recursion are inner parameters of the operator. For the \texttt{update} operator, some metaheuristics do not have an explicit \texttt{update} but directly accept $S_{new}$. In that case, \texttt{update} is the operator always updating $S$ as $S_{new}$, e.g., always accepting particles' new locations in PSO.

As presented above, the proposed prototype can couple algorithmic components from entirely different paradigms of the metaheuristic family in various structures. This enables fully discovering potentials and novelties across the family and producing high-performance algorithms that differ from what human designers have considered. All instantiations of the coherent prototype are considered to be valid. This avoids increasing too much computational complexity while allowing enough flexibility to explore many new designs.

\subsection{Graph Algorithm Representation} \label{sec_representation}
\subsubsection{Motivation}
The directed graph is a well-defined format to describe arbitrary orderings of a process. It is more flexible than the usual vector representation in expressing various algorithm structures. Cartesian GP \cite{kantschik1999meta,shirakawa2009evolution,ryser2016iterative} and the work of \cite{tisdale2021directing} have developed directed acyclic and cyclic graphs to represent metaheuristic algorithms, respectively. However, the former only expresses linear serial ordering of operators. The latter is well-developed to express various $(\mu / p + \lambda)$ and $(\mu / p,\lambda)$ EA structures but lacks generalization to describe other metaheuristics. Besides, the cyclic graph is not easy to manipulate, because it induces a bloated design space with plenty of meaningless cycles that do not connect the output forward to the beginning of the next algorithm iteration. This motivates us to develop a new acyclic graph representation that is flexible to represent various algorithm structures instantiated from the proposed prototype, and at the same time, straightforward to manipulate. 

\begin{figure}[t] 
	\centering
        \includegraphics[width=1\linewidth]{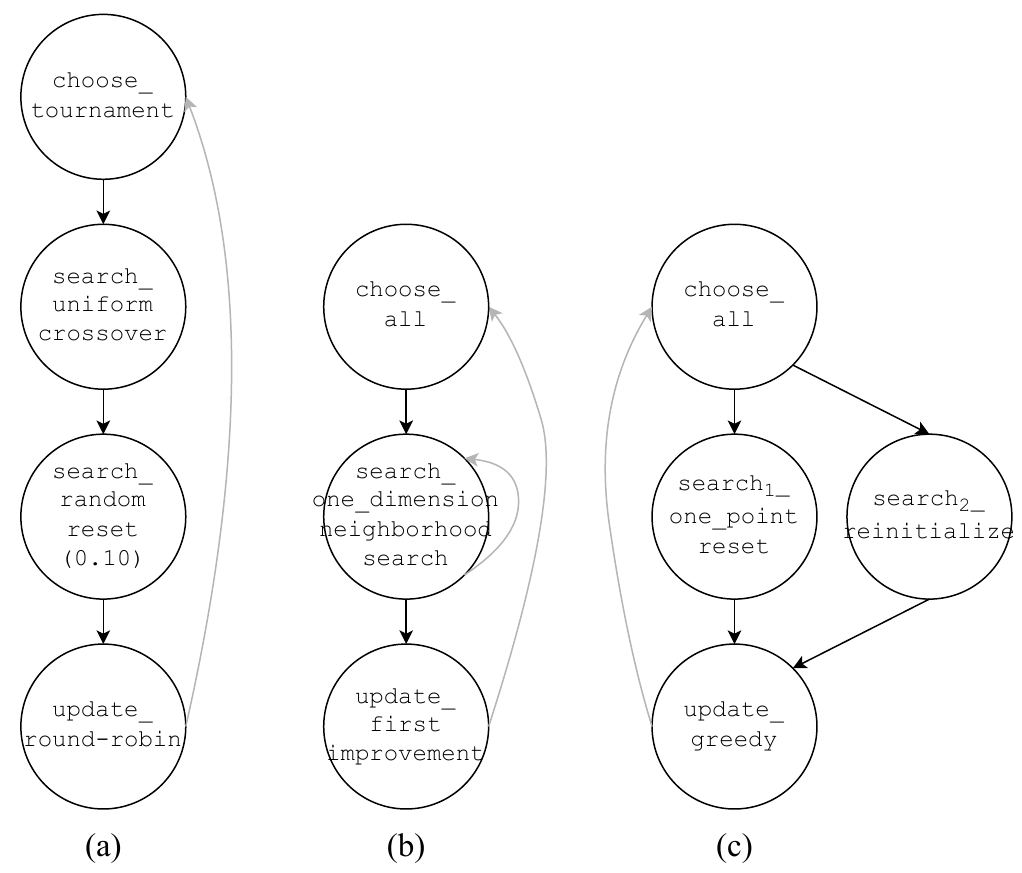}
	\caption{Three instantiations of the proposed graph algorithm representation.}
	\label{graph}
\end{figure}

Next, we illustrate the proposed representation's structure, describe its workflow, and give remarks for keeping the representation valid, with visual aid from the instantiations in Figure \ref{graph}. 

\subsubsection{Structure} 
As shown in Figure \ref{graph}, vertices of the graph represent algorithmic operators; directed edges determine the execution flow of the operators; attributes of a vertex refer to the operator's inner parameters, e.g., the reset probability of the \texttt{random reset} operator in Figure \ref{graph}(a) is $0.10$. This graph is natural to represent an algorithm at the operator level. Even so, we could find no previous work using this graph in automatically designing metaheuristic algorithms. Note that the curve edges in grey are not manipulated within the graph. Instead, they are determined by the \textbf{while} loops in the prototype of Algorithm \ref{alg_prototype}. This makes the graph itself acyclic and much easier to manipulate than cyclic ones. Each vertex is open to connecting forward to multiple vertices. This enables multiple nonlinear pathways to form various algorithm structures.

\subsubsection{Workflow} 
The execution of each vertex in the graph requires input. The initial solution(s) will be fed in, if the input of the \texttt{choose} vertex does not exist. These ensure the workflow of the graph starting from the first execution of the \texttt{choose} vertex. Then, the workflow goes along the directed edges. The output solutions of each vertex will not be evaluated until a vertex requires evaluation. For the vertex that connects forward to multiple vertices, the workflow is controlled by the \texttt{condition} in the prototype of Algorithm \ref{alg_prototype}. For example, in the iterative local search algorithm represented in Figure \ref{graph}(c), the workflow from the \texttt{choose} vertex goes to the $\texttt{search}_1$ vertex to perform local search, if $\texttt{condition}_1$ is not met; otherwise, the workflow goes to the $\texttt{search}_2$ vertex to perform global distribution. The \texttt{condition} can be predefined or evolved as an inner parameter (attribute) of its corresponding \texttt{search} operator (vertex). The workflow ends when going back to the starting vertex (\texttt{choose}) but the vertex is not allowed to execute anymore because of algorithm termination.

\subsubsection{Remarks for keeping validity} 
Since a common search pathway consists of a \texttt{choose}, \texttt{search}, and \texttt{update} operators, a valid graph should have at least three vertices. Furthermore, the graph should be connected without isolated sub-graphs. This is because isolated sub-graphs result in unrelated pathways; these pathways indicate independent fragments that do not contribute to the algorithm.

\subsection{Graph Representation Embedding} \label{sec_embedding} 
\subsubsection{Motivation}
It is natural to manipulate the graph in the adjacent list or matrix forms. However, the adjacent list is variable-length considering different algorithm structures, while the adjacent matrix is large-scale and sparse in scenarios with massive candidate design choices. The variable-length, large-scale, and sparse nature makes the graph non-trivial to be manipulated when using certain design methods or performance evaluation strategies. For example, in design scenarios with $n$ candidate choices and a surrogate model to estimate algorithms' performance, the surrogate's training sample space grows exponentially ($n^2$) if directly manipulating the graph's adjacent matrix. Such large-scale sample space greatly challenges the surrogate's accuracy, considering the limited sample availability. This motivates us to learn a compact embedding for the graph. The embedding offers an alternative form to be manipulated, ensuring AutoOpt's generality in different design scenarios. 

\subsubsection{Proposed Embedding}
\begin{figure*}[t] 
	\centering
        \includegraphics[width=0.5\linewidth]{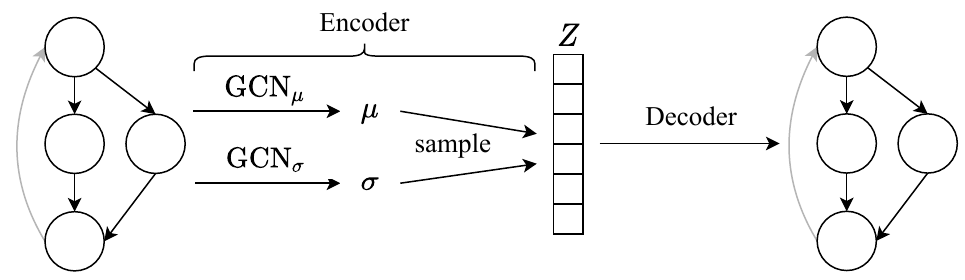}
	\caption{Illustration of the graph embedding learning process.}
	\label{embedding}
\end{figure*}

We employ the variational graph auto-encoder (VGAE) \cite{kipf2016variational} to learn the embedding because of VGAE's ability of unsupervised learning of graphs' smooth latent embedding and its well-established track records in a variety of tasks. Furthermore, although the graph contains different types of vertices, e.g., \texttt{choose}, \texttt{search}, and \texttt{update}, this does not incur heterogeneous embedding learning, and the homogeneous VGAE is sufficient, since the relations among the vertex types have been regulated by the algorithm prototype. We illustrate the process of graph representation embedding in Figure \ref{embedding}.

As shown in Figure \ref{embedding}, the input of the VGAE is the algorithm's graph representation, i.e., the adjacent matrix of the ordering of vertices (operators) and the vector of vertices' attribute values (operators' inner parameter values). For ease of manipulation, we fix the length of the attribute vector by aligning it with the vector that contains the inner parameters of all candidate operators in the design space. That means, suppose the design space has $n$ operators, and each operator has one inner parameter, we have an $n$-dimensional vector recording all the inner parameters. For a graph with vertices representing the $i$th, $j$th, and $k$th operators, we fix an $n$-dimensional attribute vector with attribute values in the $i$th, $j$th, and $k$th entities and zeroes in other entities. 

The input graph then goes into the encoding procedure. In the procedure, a graph convolutional network (the GCN in Figure \ref{embedding}) abstracts the graph's statistical information ($\mu$, $\sigma$) and uses the information to sample a compact latent embedding $Z$ for the graph. Such a sampling over the continuous latent space can learn a smoother embedding than conventional encoders. After that, $Z$ goes through the decoding procedure to reconstruct the graph. The optimal encoder and decoder are learned by minimizing the difference between the reconstructed and original graphs.

The embedding can be extracted by the learned encoder. This embedding is more compact than the original graph without much information loss due to minimizing the difference. The compact and meaningful embedding provides an alternative form of the graph to be manipulated, which benefits certain design scenarios. For example, in cases with a surrogate model to estimate algorithms' performance, the surrogate can be trained in the embedding space with much lower dimensions but richer features than the original algorithm representation space, which significantly reduces the hardness of training. The learned decoder can be used to reconstruct the graph representation from the embedding in cases with such reconstruction request.    

\subsection{Overall Framework} \label{sec_framework}  
The AutoOpt framework follows the pipeline of Figure \ref{pipeline}. The algorithm prototype proposed in subsection \ref{sec_prototype} and graph representation in \ref{sec_representation} act as prototype $\mathcal{C}$ and representation $\mathcal{R}$ in module \ding{182} of the pipeline, respectively. Their generality and flexibility enable discovering various algorithms in a single run of design, thus boosting the possibility of finding high-performance algorithms for different problems. The prototype and representation provide a general way of algorithm design at the operator level in which operators can either be formed by computational primitives or chosen from existing ones. Thus, either the design space with computational primitives or that with existing operators can cooperate with the prototype and representation. 

The embedding method proposed in \ref{sec_embedding} offers an alternative form of the graph representation to be manipulated in modules \ding{183} and \ding{184} of the pipeline. It enables producing algorithms and evaluating performance in a much more compact embedding space, which benefits design scenarios with non-trivial graph representation space caused by massive candidate choices. The proposed prototype, representation, and embedding are independent elements that can cooperate with a plenty of design space, design methods (e.g., SMAC's Bayesian optimization \cite{lindauer2022smac3}, irace's estimation of distribution \cite{lopez2016irace}, model-free local \cite{hutter2009paramils} and global \cite{koza1994genetic} search), and performance evaluation strategies (e.g., racing \cite{lopez2016irace}, intensification \cite{hutter2009paramils}, capping \cite{hutter2009paramils,de2022capping}, and surrogate model \cite{lindauer2022smac3}) to implement an algorithm design. This ensures AutoOpt's generality.

\section{Experiments and Applications}\label{sec_experiment}
In this section, we present an empirical study on AutoOpt. We first introduce the study setup. Then, we conduct an ablation analysis on the efficiency of AutoOpt's key elements via numerical benchmark problems. After that, we verify AutoOpt's overall performance and practicality by real problems. Finally, by summarizing the investigations on all problems, we observe whether the best algorithms designed for the problems are disparate. This provides evidence of AutoOpt's ability to discover various algorithm structures and the necessity of such a general framework to fit for different problem-solving. 

\subsection{Setup} \label{sec_setup}
\subsubsection{Test problems}
We adopt the $f1$ to $f10$ functions from the CEC 2005 real-valued parameter optimization problem suite \cite{suganthan2005problem} and the $f1$ to $f20$ functions from the CEC 2013 real-valued parameter optimization problem suite \cite{liang2013problem2}. They are representative continuous benchmark problems covering different Sphere, Schwefel, Elliptic, Rosenbrock, Griewank, Ackley, and Rastrigin  versions etc. We set each function as an independent targeted problem with three instances. These instances differ in their dimensions of solution space, i.e., $10-$, $30-$, and $50-D$, respectively. We further utilize two real problems. The first is a power system restoration scheduling problem with highly constrained binary solution space. The second is a discrete non-separable beamforming problem from a reconfigurable intelligent surface (RIS)-aided communication system. Each problem has a number of instances that differ in their input variable values. Metaheuristics could meet the discrete, highly-constrained, and non-separable challenges. In this regard, the automated design could benefit the problem researchers to quickly get an efficient metaheuristic solver from the variety of choices.

\subsubsection{Baselines} 
We use different types of algorithms across the metaheuristic family as baselines to make a comprehensive empirical comparison. They include covariance matrix adaptation evolution strategy (CMA-ES) \cite{wierstra2014natural}, EDA \cite{larranaga2001estimation}, differential evolution (DE) \cite{price2013differential}, PSO \cite{shi1998modified}, bi-population CMA-ES (BIPOP) \cite{loshchilov2013cma}, and success history-based adaptive DE (SHADE) \cite{tanabe2013success} for continuous problems, and GA \cite{holland1973genetic}, ILS \cite{lourencco2003iterated}, and SA \cite{kirkpatrick1983optimization} for discrete problems. We also adopt CPLEX \footnote{\url{https://www.ibm.com/products/ilog-cplex-optimization-studio/cplex-optimizer}} as a mathematical solver baseline in the real applications to verify AutoOpt's practicality.

\subsubsection{Implementation details} \label{implementation}
We implement AutoOpt by using the design space with operators shown in Table \ref{operator}, ParamILS's model-free search \cite{hutter2009paramils} as the design method, and intensification \cite{hutter2009paramils} as the performance evaluation strategy. Since AutoOpt's key elements are independent of the design space, design method, and performance evaluation strategy, a particular implementation will not bias the ablation analysis. We set the execution of the implementation as follows: \textit{i) Targeted problem}: For each numerical benchmark problem, two of its instances are randomly selected and targeted during algorithm design; the remaining instance is for testing and comparing with the baselines. For each real problem, half of its instances are for design; the remaining half is for test and comparison. \textit{ii) Computational effort}: For each targeted problem, AutoOpt produces $10$ algorithms at each iteration (i.e., the iteration of modules \ding{183} and \ding{184} in Figure \ref{pipeline}) and terminates after evaluating 5000 produced algorithms. In each evaluation, the algorithm runs ten times on each targeted problem instance; 5000 function evaluations are granted for each run. For a fair comparison with the baselines, all the produced algorithms conduct population-based search with a population size of $20$ during evaluation. We set the maximum number of operators in each search pathway as $6$ to avoid unnecessarily bloated algorithms. \textit{iii) Performance measurement}: Algorithm's performance on a targeted problem is measured by the average solution fitness over all runs on all instances. The algorithm with the best performance found during the design will be the final algorithm and will be further compared with the baselines. 
\begin{table*}[t]
\centering
\caption{Algorithmic operators within the implemented design space.}
\label{operator}
\begin{tabular}{ll}
\toprule
Operators & Description \\
\midrule
\texttt{Choose}:                      & \\
\texttt{choose\_traverse}             & Choose each of the current solutions to search from \\
\texttt{choose\_roulette\_wheel}      & Roulette wheel selection \\
\texttt{choose\_tournament}           & \textit{K}-tournament selection \\
\texttt{choose\_cluster}              & Brain storm optimization's idea picking up for choosing solutions to search from \cite{shi2015optimization} \\
\midrule
\texttt{Search} (continuous):         & \\
\texttt{cross\_arithmetic}            & Whole arithmetic crossover \\
\texttt{cross\_sim\_binary}           & Simulated binary crossover \cite{deb1995simulated} \\
\texttt{cross\_point\_one}            & One-point crossover \\
\texttt{cross\_point\_two}            & Two-point crossover \\
\texttt{cross\_point\_n}              & \textit{n}-point crossover \\
\texttt{cross\_point\_uniform}        & Uniform crossover \\
\texttt{search\_cma}                  & The evolution strategy with covariance matrix adaption \\
\texttt{search\_eda}                  & The estimation of distribution \\
\texttt{search\_mu\_cauchy}           & Cauchy mutation \cite{yao1999evolutionary} \\
\texttt{search\_mu\_gaussian}         & Gaussian mutation \cite{fogel1998artificial} \\
\texttt{search\_mu\_polynomial}       & Polynomial mutation \cite{deb1996combined} \\
\texttt{search\_mu\_uniform}          & Uniform mutation  \\
\texttt{search\_pso}                  & Particle swarm optimization's particle fly and update \cite{shi1998modified} \\
\texttt{search\_de\_random}           & The \textit{random/1} differential mutation \cite{storn1997differential} \\
\texttt{search\_de\_current}          & The \textit{current/1} differential mutation \\
\texttt{search\_de\_current\_best}    & The \textit{current-to-best/1} differential mutation \\
\texttt{reinit\_continuous}           & Random reinitialization for continuous problems \\
\midrule                             
\texttt{search} (discrete):           &  \\
\texttt{cross\_point\_one}            & One-point crossover \\
\texttt{cross\_point\_two}            & Two-point crossover \\
\texttt{cross\_point\_n}              & \textit{n}-point crossover \\
\texttt{cross\_point\_uniform}        & Uniform crossover \\
\texttt{search\_reset\_one}           & Reset a randomly selected entity to a random value \\
\texttt{search\_reset\_rand}          & Reset each entity to a random value with a probability \\
\texttt{search\_reset\_creep}         & Add a small positive or negative value to each entity with a probability, for problems with ordinal attributes \\
\texttt{reinit\_discrete}             & Random reinitialization for discrete problems \\ 
\midrule
\texttt{Update}:                      & \\
\texttt{update\_always}               & Always select new solutions \\
\texttt{update\_greedy}               & Select the best solutions \\
\texttt{update\_pairwise}             & Select the better solution from each pair of old and new solutions \\
\texttt{update\_round\_robin}         & Select solutions by round-robin tournament \\
\texttt{update\_simulated\_annealing} & Simulated annealing's update mechanism, i.e., accept worse solution with a probability \cite{kirkpatrick1983optimization}    \\
\bottomrule
\end{tabular}
\end{table*}

For the baselines, DE is in the \textit{current/1} manner \cite{price2013differential}; GA is with the one-point crossover and random reset mutation; ILS and SA employ the one-dimensional random reset as the search operator; CMA-ES, EDA, PSO, BIPOP, and SHADE are the default versions from \cite{wierstra2014natural,larranaga2001estimation,shi1998modified,loshchilov2013cma,tanabe2013success}, respectively. DE and PSO use parameter settings from \cite{price2013differential} and \cite{shi1998modified}, respectively. The source code of implementations of AutoOpt and the baselines is publicly available \footnote{\url{https://github.com/auto4opt/AutoOpt}}. In the peer comparison, algorithms' (the final algorithm returned by AutoOpt and the baselines) population size is set to 50; algorithms terminate after $1000*D$ function evaluations on the benchmark problems and 50000 function evaluations on the real problems; each algorithm runs 30 times on each problem.

\subsection{Ablation Analysis}
AutoOpt has three key elements, i.e., the general algorithm prototype, graph algorithm representation, and graph representation embedding. While in subsections \ref{sec_prototype} and \ref{sec_representation}, we have clarified how AutoOpt can express and discover various algorithm structures via the prototype and representation, the efficiency of embedding requires empirical ablation analysis to verify. To this end, we create an algorithm design scenario in which the performance evaluation strategy intensification \cite{hutter2009paramils} in the AutoOpt implementation is replaced with a random forest (RF) surrogate model \cite{lindauer2022smac3}. The surrogate's input is a produced algorithm; the output is its estimated performance. To train the surrogate, one should sample a number of algorithms (training data) over the design space and exactly evaluate their performance (data labels) through running the algorithm on the targeted problem. This incurs a few-shot training task due to the heavy and limited computational load in labelling the data via exact performance evaluation. In such a scenario, an eligible embedding is assumed to ease the few-shot training and in turn benefits the algorithm design. 

For each targeted problem, we randomly sample $1000$ algorithms to train the surrogate model. The trained model is then used to estimate the performance of algorithms produced during design. We derive two versions, one with the surrogate model trained and used in the original graph representation space and the other with the surrogate model trained and used in the embedding space. The original space in forms of the graph's adjacent matrix and vertices' attribute vector is over $600$ dimensions, while the embedding space is reduced to $20$ dimensions by VGAE. We investigate the embedding's efficiency by 1) comparing the estimation accuracy of the surrogate trained in the embedding space with that trained in the original space, and 2) comparing the performance of the algorithm designed with embedding with that designed without embedding and baselines.

\subsubsection{Accuracy of the surrogate with embedding}
The surrogate's accuracy on $f1$ to $f10$ of the CEC 2005 problems is shown in Table \ref{surrogate_accuracy}. As the surrogate works on identifying better algorithms among current ones, its accuracy indicates to what extent the estimated performance referring a correct algorithm performance rank. Thus, we measure the accuracy by Kendall's $\tau$ correlation coefficient \cite{kendall1948rank} between algorithms' rank in terms of the estimated performance and that in terms of the exactly evaluated performance. The measurement is conducted on the algorithms produced during design, which are separate from those for training the surrogate. Higher $\tau$ values (with a maximum $1$) indicate better accuracy. From Table \ref{surrogate_accuracy}, the surrogate trained in the embedding space (RF\_embed) constantly obtains better accuracy than that trained in the original graph representation space (RF\_no\_embed) on all problems. This observation demonstrates that the embedding is meaningful, and the much more compact embedding space significantly eases the surrogate training to get higher accuracy.  
\begin{table*}[t]
\centering
\caption{Efficiency of the embedding in training performance estimation surrogate}
\label{surrogate_accuracy}
\begin{tabular}{lccccccccccc}
\toprule
              & $f1$   & $f2$   & $f3$   & $f4$   & $f5$   & $f6$   & $f7$   & $f8$   & $f9$   & $f10$  & Average \\ 
\midrule
RF\_embed     & 0.8111 & 0.7941 & 0.7785 & 0.7930 & 0.7912 & 0.7567 & 0.7980 & 0.6807 & 0.8015 & 0.8196 & 0.7824  \\
RF\_no\_embed & 0.7694 & 0.7779 & 0.7641 & 0.7916 & 0.7809 & 0.7372 & 0.7697 & 0.6729 & 0.7951 & 0.8164 & 0.7675  \\
MLP           & 0.7159 & 0.7689 & 0.5414 & 0.7725 & 0.7985 & 0.5250 & 0.8699 & 0.5045 & 0.8257 & 0.8605 & 0.7183  \\
RBF           & 0.5985 & 0.5489 & 0.6194 & 0.5776 & 0.6088 & 0.5756 & 0.5608 & 0.5162 & 0.5668 & 0.5903 & 0.5763  \\
\bottomrule
\end{tabular}
\end{table*} 

We further compare with two neural network surrogates, i.e., multilayer perception (MLP) and radial basis function (RBF) networks \footnote{As a usual practice, the MLP has two fully connected hidden layers both with the ReLU activation function; each hidden layer reduces half of the size of its input; the learning rate is set as $0.005$. The RBF has one hidden layer with the RBF activation function.}. Both networks are trained in the original graph representation space. Although neural networks are recognized in their nonlinear expressiveness and inherently perform embedding, the results in Table \ref{surrogate_accuracy} show that MLP and RBF are overall inferior to RF\_embed due to the limited availability of training data. These results again confirm the efficiency of the proposed embedding in enhancing the surrogate in the few-shot training task and the necessity of the embedding-based RF surrogate in algorithm design. 

\subsubsection{Performance of the algorithm designed with embedding}
In this subsection, we first ablate the proposed graph embedding through the CEC 2005 problems. The average and standard deviation of the performance over the 30 runs on the problems is shown in Table \ref{alg_performance}. In Table \ref{alg_performance}, AutoOpt\_embed refers to the algorithms designed by the AutoOpt version with a performance estimation surrogate trained in the embedding space; AutoOpt\_no\_embed refers to the algorithms designed by the AutoOpt version with a performance estimation surrogate trained in the original space. We then investigate the performance of the complete AutoOpt (with the embedding) by comparing with the state-of-the-art BIPOP, SHADE as well as the CMA-ES, EDA, PSO, and DE baselines on the more challenging CEC 2013 benchmarks. The comparison results are given in Table \ref{alg_performance2}. As mentioned in the setup of subsection \ref{sec_setup}, we adopt AutoOpt to design algorithms for each problem separately, which means AutoOpt\_embed refers to different designed algorithms for different problems, so does AutoOpt\_no\_embed. Besides, the performance is obtained on the test instances independent from those targeted during algorithm design. 

According to Table \ref{alg_performance}, AutoOpt\_embed together with AutoOpt\_no\_embed win the comparisons with baselines on 9 out of the 10 problems. This result verifies AutoOpt's ability to discover eligible algorithms. In comparing AutoOpt\_embed and AutoOpt\_no\_embed, we observe that AutoOpt\_embed is better than AutoOpt\_no\_embed on 7 problems and is comparable with AutoOpt\_no\_embed on 2 out of the remaining 3 problems. The better results attribute to the more accurate surrogate trained in the embedding space. This indicates that the more accurate identification of promising algorithms benefits guiding the design toward finding more efficient algorithms. 

Further comparisons on the CEC 2013 problems (Table \ref{alg_performance2}) reveal a surprising finding: AutoOpt demonstrates performance comparable to the CEC 2013 competition winner BIPOP, achieving better results on 7 out of 20 problems compared to BIPOP's 9 wins, and significantly outperforms the other state-of-the-art SHADE. This highlights AutoOpt's promising in algorithm design, particularly given that BIPOP and SHADE are meticulously crafted by experts with advanced adaptive strategies, while the algorithms designed by AutoOpt rely solely on the general components listed in Table \ref{operator}.

Moreover, neither BIPOP nor SHADE consistently outperforms across all problems, and the performance rankings of the four baselines remain relatively balanced. These findings underscore the limitation of fixed algorithm structures for diverse problem-solving scenarios. AutoOpt's ability to tailor algorithms, therefore, represents a significant advancement toward generalized, high-performance optimization.
\begin{table*}[t]
\centering
\caption{Average and standard deviation of algorithms' performance on the CEC 2005 problems. Best results according to Wilcoxon sign test are in bold.}
\scriptsize
\label{alg_performance}
\begin{tabular}{lrrrrrr}
\toprule
      & AutoOpt\_embed           & AutoOpt\_no\_embed       & CMA-ES           & EDA              & PSO               & DE                \\
\midrule
$f1$  &     \textbf{-449.95$\pm$5.60E-02} &     \textbf{-449.70$\pm$8.38E-01} &      491.69$\pm$3.27E+02 &     \textbf{-449.96$\pm$1.93E-03} &     38315.50$\pm$7.37E+03 &    18054.24$\pm$3.40E+03 \\
$f2$  &     \textbf{1673.64$\pm$1.01E+03} &    2198.51$\pm$3.02E+03 &     6039.79$\pm$1.24E+03 &    3827.35$\pm$1.28E+03 &     36490.28$\pm$6.39E+03 &    36117.86$\pm$6.57E+03 \\
$f3$  & \textbf{5444596.87$\pm$1.93E+06} & 62163892.97$\pm$9.64E+06 & 31318518.16$\pm$7.35E+06 & 40496812.27$\pm$4.23E+06 &  97193764.51$\pm$1.39E+08 & 92120350.84$\pm$2.17E+07 \\
$f4$  &    \textbf{5435.44}$\pm$5.88E+03 &    5719.35$\pm$3.37E+03 &    6661.41$\pm$6.15E+03 &    6891.51$\pm$1.89E+03 &     43523.30$\pm$6.54E+03 &    57290.99$\pm$9.09E+03 \\
$f5$  &     3741.07$\pm$4.20E+02 &     3713.09$\pm$2.35E+02 &     \textbf{3163.59$\pm$1.23E+03} &     3359.79$\pm$2.59E+02 &     23757.44$\pm$3.89E+03 &    10437.74$\pm$1.12E+03 \\
$f6$  &    \textbf{73630.65$\pm$2.27E+04} & 78998.70$\pm$2.90E+04 & 17477444.02$\pm$1.40E+07 &   154174.95$\pm$1.23E+05 &  85983509.43$\pm$3.52E+09 & 31141442.98$\pm$1.01E+09 \\
$f7$  &     -163.31$\pm$2.96E+00 &     -159.83$\pm$1.22E+01 &     -120.64$\pm$1.23E+01 &     \textbf{-170.20$\pm$3.07E+00} &      1424.08$\pm$3.93E+02 &      561.86$\pm$1.07E+02 \\
$f8$  &     \textbf{-119.15$\pm$1.78E-01} &     \textbf{-119.10$\pm$6.12E-02} &     \textbf{-119.01$\pm$8.20E-02} &     \textbf{-119.00$\pm$4.23E-02} &      \textbf{-119.15$\pm$1.05E-01} &     \textbf{-118.99$\pm$8.94E-02} \\
$f9$  &     \textbf{-313.45$\pm$3.51E+00} & -301.98$\pm$5.92E+00 &     -156.59$\pm$1.10E+01 &     -145.77$\pm$1.36E+01 &       -38.42$\pm$2.24E+01 &      -48.12$\pm$1.97E+01 \\
$f10$ &     -214.32$\pm$1.96E+01 &     \textbf{-218.13$\pm$2.89E+01} &     -115.47$\pm$7.12E+00 &     -138.37$\pm$8.44E+00 &        118.05$\pm$5.26E+01 &      226.18$\pm$7.27E+01 \\
\bottomrule
\end{tabular}
\end{table*}

\begin{table*}[t]
\centering
\caption{Average and standard deviation of algorithms' performance on the CEC 2013 problems. Best results according to Wilcoxon sign test are in bold.}
\tiny
\label{alg_performance2}
\begin{tabular}{lrrrrrrrr}
\toprule
      & AutoOpt\_embed           & BIPOP    & SHADE   & CMA-ES           & EDA              & PSO               & DE                \\
\midrule
f1  & -1352.94±4.86E+00     &  \textbf{-1400±4.29E+01}     & -1398.59±4.28E+01       & -485.61±7.85E+02        & -1397.43±4.06E+00      & 137698.68±1.70E+04      & -1013.23±6.76E+01       \\ 
f2  & 17542072.16±5.44E+06  &  \textbf{123933.94±4.57E+01} & 176629134.38±6.31E+01   & 16888864.38±4.56E+06    & 16870706.39±2.04E+06   & 2125701659.64±6.35E+08  & 266580377.27±5.97E+07   \\ 
f3  & 159892888.77±1.51E+07 &  \textbf{324964.46±4.21E+01} & 72460383626.01±6.08E+01 & 46729754692.49±3.08E+10 & 3875874241.85±2.25E+09 & 36948258581618±1.56E+13 & 75570914990.71±1.27E+10 \\
f4  & 77912.48±1.23E+04     &  \textbf{-1041.24±4.27E+01}  & 161960.72±6.03E+01      & 209168.62±2.22E+04      & 72770.28±1.90E+04      & 376068.53±9.93E+04      & 196883.51±1.09E+04      \\
f5  & -980.59±2.63E+00      & \textbf{-1000±4.29E+01}     & -998.38±4.28E+01        & -321.96±9.22E+02        & -286.12±8.07E+02       & 147319.18±5.29E+04      & -877.68±9.10E+00        \\ 
f6  &  \textbf{-876.07±3.73E+00}      & -856.55±4.67E+01   & -806.59±4.67E+01        & 283.07±6.45E+02         & -769.3±6.11E+01        & 49065.84±1.06E+04       & 231.62±1.64E+02         \\
f7  & -741.08±1.10E+01      &  \textbf{-798.56±4.09E+01}   & -601.17±6.54E+01        & -662.87±1.95E+01        & -733.32±1.43E+01       & 3119.96±1.35E+03        & -505.8±1.78E+01         \\
f8  &  \textbf{-678.83±3.63E-02}      & -678.77±4.36E+01   & -678.8±6.27E+01         & -678.79±3.37E-02        & -678.81±6.01E-02       & -678.73±2.87E-02        & -678.78±1.43E-02        \\
f9  &  \textbf{-588.08±3.76E+00}      & -579.96±5.16E+01   & -538.2±6.25E+01         & -586.25±3.03E+00        & -585.38±4.26E+00       & -525.19±1.15E+00        & -540.06±2.03E+00        \\
f10 & -374.49±7.10E+01      &  \textbf{-499.99±4.29E+01}   & 555.07±4.98E+01         & -221.69±1.94E+02        & -317.81±6.49E+01       & 18109.67±2.59E+03       & 1974.3±2.56E+02  \\
f11 &  \textbf{-396.81±8.37E-01} & -367.17±4.18E+01 & 555.07±4.45E+01   & -335.47±3.16E+01 & -388.58±3.35E+00  & 2873.64±6.73E+02    & -219.54±5.49E+00  \\ 
f12 & 75.79±1.16E+01   & -260.2±4.27E+01  &  \textbf{-364.51±5.40E+01}  & -100.09±1.72E+02 & 54.97±1.15E+01    & 3867.1±1.03E+03     & 460.15±5.06E+01   \\ 
f13 & -28.72±1.22E+02  &  \textbf{-94.37±4.24E+01}  & 333.28±5.26E+01   & 96.69±1.20E+02   & 154.73±2.00E+01   & 3380.65±1.01E+03    & 593.55±5.79E+01   \\
f14 &  \textbf{-62.86±8.48E+00}  & 2063.19±4.59E+01 & 548.54±5.66E+01   & 7740.07±3.33E+03 & 13483.01±5.22E+02 & 16327.88±5.83E+02   & 4792.69±4.72E+02  \\
f15 & 7469.67±4.93E+02 & 3341.71±5.42E+01 &  \textbf{1839.21±6.25E+01}  & 5516.42±4.35E+03 & 13513.78±2.22E+02 & 14909.77±5.24E+02   & 13561.57±4.17E+02 \\ 
f16 &  \textbf{201.84±2.72E-01}  & 203.88±5.64E+01  & 12965.04±6.19E+01 & 204±3.34E-01     & 203.92±3.69E-01   & 204.68±9.55E-01     & 203.94±3.42E-01   \\
f17 & 312.9±1.43E-01   & 384.91±2.57E+01  &  \textbf{203.92±2.59E+01}   & 389.69±6.53E+01  & 637.43±1.80E+01   & 5748.53±2.30E+02    & 595.48±1.66E+01   \\
f18 &  \textbf{413.18±2.24E-01}  & 796.62±2.55E+01  & 418.92±3.06E+01   & 514.01±1.55E+02  & 728.91±1.52E+01   & 5929.84±8.24E+02    & 702.15±1.01E+01   \\
f19 & 539.66±3.15E-01  &  \textbf{504.71±4.21E+01}  & 1083.5±4.38E+01   & 547.47±1.73E+01  & 545±5.22E-01      & 15366052.5±5.53E+06 & 2088.62±3.47E+02  \\
f20 & 624.17±1.79E-01  & 621.29±4.41E+01  &  \textbf{514.06±7.54E+01}   & 623.7±3.87E-01   & 624.24±9.52E-02   & 624.58±6.25E+02     & 624.49±5.98E-02  \\
\bottomrule
\end{tabular}
\end{table*}

Finally, we report three representatives (in terms of algorithm structure) among the designed algorithms in Algorithms \ref{alg_f1}, \ref{alg_f2}, and \ref{alg_f9} \footnote{For brevity, the \textbf{while} loop for assigning computational effort to the enclosed search pathway (i.e., line 6 of Algorithm \ref{alg_prototype}) is omitted, if the enclosed search pathway executes only once without looping. Values in brackets refer to the operator's inner parameter values, e.g., the crossover probability is $0.21$ in line 3 of Algorithm \ref{alg_f2}}. Among them, Algorithms \ref{alg_f1} and \ref{alg_f2} have one search pathway with 3 and 4 operators in series, respectively; Algorithm \ref{alg_f9} contains an inner loop of the uniform mutation operator that performs iterative local refinement via a small mutation probability of $0.081$. Although the fittest algorithm for each benchmark problem may not be unique, the designed algorithms clearly demonstrate AutoOpt's efficiency in discovering diverse algorithm structures.
\begin{algorithm}[p]
\caption{Designed algorithm for CEC 2005 $f1$} 
\label{alg_f1}
\footnotesize
\KwIn{initial solutions $S$}
\While{algorithm not terminate}{		
    $S=\texttt{choose\_traverse}(S)$ \\					
    $S_{new}=\texttt{search\_eda}(S)$ \\
    $S=\texttt{update\_greedy}(S,S_{new})$ \\		
}
\KwOut{best solution from $S$}
\end{algorithm}

\begin{algorithm}[p]
\caption{Designed algorithm for CEC 2005 $f2$} 
\label{alg_f2}
\footnotesize
\KwIn{initial solutions $S$}
\While{algorithm not terminate}{		
    $S=\texttt{choose\_tournament}(S)$ \\			
    $S_{new}=\texttt{cross\_arithmetric}(0.21,S)$ \\			
    $S_{new}=\texttt{search\_mu\_polynomial}(0.23,25.68,S_{new})$ \\
    $S=\texttt{update\_pairwise}(S,S_{new})$ \\		
}
\KwOut{best solution from $S$}
\end{algorithm}

\begin{algorithm}[p]
\caption{Designed algorithm for CEC 2005 $f9$} 
\label{alg_f9}
\footnotesize
\KwIn{initial solutions $S$}
\While{algorithm not terminate}{		
    $S=\texttt{choose\_traverse}(S)$ \\			
    $S_{new}=\texttt{search\_mu\_polynomial}(0.19,33.03,S)$ \\
    $S=\texttt{update\_pairwise}(S,S_{new})$ \\
    \While{``consume $500$ function evaluations" not met}{
    $S=\texttt{choose\_traverse}(S)$ \\			
    $S_{new}=\texttt{search\_mu\_uniform}(0.081,S)$ \\
    $S=\texttt{update\_pairwise}(S,S_{new})$ \\
    }
}
\KwOut{best solution from $S$}
\end{algorithm}

\subsection{Application to Power System Restoration}
\subsubsection{Problem Description}
Power system restoration refers to scheduling the black start (BS) and non-black start (NBS) units to re-energize the power network and loads after the power grid completely blackouts. A high-quality restoration solution is significant to ensure the power system's safety and stability.

The targeted power system restoration problem considers a restoration process as follows. First, BS units restart. Then, NBS units reboot by receiving cranking power from BS units. During rebooting, some critical loads are re-energized to maintain the frequency and voltage of the power system. Finally, after all NBS units have been rebooted, the remaining loads are re-energized according to their priority. Decision variables of the problem are the restoration sequence and path of the BS units, NBS units, and loads; the objective is to minimize the total restoration time:
\begin{subequations}
	\label{eq_restore}
	\begin{align}
        &\min_{\mathbf{v}_a,\mathbf{h}_b}\quad\sum_{a=1}^{A}\mathbf{v}_{a}^{T}\mathbf{w}_{a}+\sum_{b=1}^{B}\mathbf{h}_{b}^{T}\mathbf{l}_{b}, \\ 
        &\quad s.t.\quad \sum_{c=1}^{C}v_{ca}=1, a=1,2,\cdots,A, \\ 
        &\qquad\quad\    \sum_{a=1}^{A}h_{ab}=1, b=1,2,\cdots,B, \\
        &\qquad\quad\    0\leq P_{i}^{gen}\leq P_{i}^{gen,max}, \forall i\in\Omega_{Gen}, \\
        &\qquad\quad\    Q_{i}^{gen,min}\leq Q_{i}^{gen}\leq Q_{i}^{gen,max}, \forall i\in\Omega_{Gen}, \\
        &\qquad\quad\    0\leq P_{a,t-1}^{gen}-P_{a,t}^{gen}\leq R_{G}\Delta t, \forall a\in\Omega_{NBS}, \\
        &\qquad\quad\    P_{m,t}=\sum_{mn\in\Omega_{Br}}P_{mn,t}, \\
        &\qquad\quad\    Q_{m,t}=\sum_{mn\in\Omega_{Br}}Q_{mn,t}, \\
        &\qquad\quad\    P_{mn}^{2}+Q_{mn}^{2}\leq S_{mn}^{2}z_{mn}, \forall mn\in\Omega_{Br}, \\
        &\qquad\quad\    U_{m}^{min}\leq U_{m}\leq U_{m}^{max}, \forall m\in\Omega_{Bus}, \\
        &\qquad\quad\    \sum_{t=1}^{T}(1-z_{mc,t})\geq t_{c}^{start}, \forall mc\in\Omega_{Br\_BS},\forall c\in\Omega_{BS}, \\
        &\qquad\quad\    z_{ma,t}\leq z_{a,t}\leq\sum_{ma\in\Omega_{Br\_NBS}}z_{ma,t}, \forall a\in\Omega_{NBS},
	\end{align}
\end{subequations}
where binary vectors $\mathbf{v}_{a}=(v_{1a},\cdots,v_{ca},\cdots,v_{Ca})$, $a=1,2,\cdots,A$, and $\mathbf{h}_{b}=(h_{1b},\cdots,h_{ab},\cdots,h_{Ab})$, $b=1,2,\cdots,B$, are decision variables that determine the connectivity from BS units to NBS units and that from NBS units to loads, respectively; $A$, $B$, and $C$ are the numbers of NBS units, loads, and BS units, respectively; $\mathbf{v_{a}}$ donates which BS unit supplies cranking power to NBS unit $a$; $\mathbf{h_{b}}$ determines which NBS unit supplies electricity to load $b$; $\mathbf{w_{a}}$ and $\mathbf{l_{b}}$ stand for the distances of the shortest time path from BS units to NBS units and that from NBS units to loads, respectively; $P_{i}^{gen}$, $P_{m,t}$, and $P_{mn}$ are the active power of generator $i$, bus $m$, and branch $mn$, respectively; $Q_{i}^{gen}$, $Q_{m,t}$ and $Q_{mn}$ are the reactive power of generator $i$, bus $m$, and branch $mn$, respectively; $P_{i}^{gen,max}$ is the maximum active power of generator $i$; $Q_{i}^{gen,min}$ and $Q_{i}^{gen,max}$ are the minimum and maximum reactive power of generator $i$, respectively; $R_{G}$ is generators' ramp rate; $S_{mn}$ is the apparent power of branch $mn$; $U_{m}$ is the voltage of bus $m$; $U_{m}^{min}$ and $U_{m}^{max}$ are the minimum and maximum voltage of bus $m$, respectively; $t_{c}^{start}$ is the time that BS unit $c$ starts generating power; $z_{mn}$, $z_{mc,t}$, $z_{ma,t}$, and $z_{a,t}$ are the restore status of branch $mn$, that of branch $mc$ that directly connects to BS unit $c$ at time $t$, that of branch $ma$ that directly connects to NBS unit $a$ at time $t$, and that of NBS unit $a$ at time $t$, respectively.

Equation \eqref{eq_restore} refers a binary integer programming problem with highly non-convex constraints, which is generally NP-hard. Finding a feasible solution that satisfies the non-convex constraints by canonical mathematical programming solvers is tricky. Besides, mathematical solvers tend to incur computational overhead when the power system scales up. Metaheuristics could meet these challenges, while AutoOpt would provide the problem researchers quick access to an efficient metaheuristic solver from a variety of choices. 

\subsubsection{Applying AutoOpt} 
We apply AutoOpt to design metaheuristic solvers for the power system restoration problem in \eqref{eq_restore}. AutoOpt is set according to the setup in subsection \ref{sec_setup}. We implement two problem instances of the IEEE 39-bus system; the instances are differ in the locations of BS and NBS units. One instance is targeted during algorithm design; the other instance is adopted for comparing the designed algorithm with baselines. 

Pseudocode of the designed algorithm (Alg$_{\rm restor}$) is given in Algorithm \ref{alg_restoration}. Constraints are handled by penalizing solutions' fitness with constraint violations. Alg$_{\rm restor}$ is a GA-style algorithm, which supports the strong track records of GA's efficient search ability on discrete problem space \cite{whitley2019next}. Specifically, the uniform crossover with a crossover rate of $0.34$ exchanges restoration path segments between a pair of promising (determined by the tournament mating selection) solutions; the one-dimensional reset, followed by the pair-wise selection, perform a relatively steady exploitation over the constrained search space. 
\begin{algorithm}[t]
\caption{Alg$_{\rm restor}$} 
\label{alg_restoration}
\footnotesize
\KwIn{initial solutions $S$}
\While{stopping criterion not met}{		
    $S=\texttt{choose\_tournament}(S)$ \\			
    $S_{new}=\texttt{cross\_point\_uniform}(0.34,S)$ \\			
    $S_{new}=\texttt{search\_reset\_one}(S_{new})$ \\
    $S=\texttt{update\_pairwise}(S,S_{new})$ \\		
}
\KwOut{best solution from $S$}
\end{algorithm}

We investigate the efficiency of the designed algorithm by comparing with the mathematical programming baseline CPLEX and two metaheuristic baselines, GA and ILS. Algorithms' performance is reported in Table \ref{tab_restoration}. The performance is measured by final solutions' fitness, i.e., the restoration time. From Table \ref{tab_restoration}, CPLEX is inferior to metaheuristic solvers, because CPLEX requires to relax the fragmented search space to be smooth, which hinders it to reach the true optimal. Alg$_{\rm restor}$ wins the best among the metaheuristic peers. This result demonstrates Alg$_{\rm restor}$'s efficiency and confirms AutoOpt's validity in tailoring efficient solvers to hard problems.  

\begin{table}[t]
\centering
\caption{Average and standard deviation of performance on power system restoration. Best results are in bold.}
\label{tab_restoration}
\begin{tabular}{p{4cm}p{4cm}}
\toprule
Algorithms                & Restoration Time      \\ 
\midrule
Alg$_{\rm restor}$ & \textbf{3030.9$\pm$2.77E+01} \\     
GA                        & 3194.8$\pm$2.38E+02   \\
ILS                       & 3268.5$\pm$1.43E+03   \\
CPLEX                     & 3697.8$\pm$0.00E+00   \\
\bottomrule
\end{tabular}
\end{table}

\subsection{Application to Beamforming in RIS-aided Communications}
\subsubsection{Problem Description} 
Reconfigurable intelligent surface (RIS) is an emerging technology for cost-effective communications \cite{yuan2020intelligent}. It is a planar passive radio structure with reconfigurable passive elements. Each element can independently adjust the phase shift on the incident signal. These elements collaboratively yield a directional beam to enhance the received signal's quality. 

We consider the RIS-aided downlink multi-user multiple-input single-output system from \cite{yan2022fitness}. In the system, a base station (BS) equipped with multiple antennas transmits signals to K single-antenna users; a RIS with a number of $N$ elements is deployed between the BS and users to provide non-line-of-sight links. The target is to  maximize the sum rate of all users subject to the transmit power constraint, by jointly optimizing the continuous active beamforming of BS and discrete phase shifts of RIS \cite{yan2022fitness}:
\begin{subequations}
	\label{eq-maxSumRate}
	\begin{align}
        &\max_{\wt_k, \mTa}\quad \sum_{k=1}^{K}\log _{2}(1+\frac{|(\bht_{\fd,k}^\mathrm{H}+\bht_{\fr,k}^\mathrm{H}\mTa\Gt)\wt_k|^2}{\sum_{j\neq k}^K| (\bht_{\fd,k}^\mathrm{H}+\bht_{\fr,k}^\mathrm{H}\mTa\Gt)\wt_{j}|^{2}+\sigma ^{2}}),\tag{\ref{eq-maxSumRate}{a}}\label{eq-Pa} \\ 
        &\quad s.t.\quad \theta_n=\beta_ne^{j\phi_n},\tag{\ref{eq-maxSumRate}{b}}\label{eq-Pb} \\ 
        &\qquad\quad\ \phi_n=\frac{\tau_n2\pi}{2^b}, \tau_n\in\{0,...,2^b-1\},\tag{\ref{eq-maxSumRate}{c}}\label{eq-Pc} \\
        &\qquad\quad\ \sum_{k=1}^{K}\|\mathbf{w}_k\|^{2}\leq P_{T},\tag{\ref{eq-maxSumRate}{d}}\label{eq-Pd}		
	\end{align}
\end{subequations}
where $\wt_k\in\bbC^{M\times 1}$ is the active beamforming at the BS towards user $k$; $\mTa=diag(\theta_1,...,\theta_n,...,\theta_N)$ is a diagonal matrix with RIS phase-shifts being the diagonal values; $\bht_{\fd,k}\in\bbC^{M\times1}$, $\Gt\in\bbC^{N\times M}$, and $\bht_{\fr,k}\in\bbC^{N\times1}$ are the channels BS-user $k$, BS-RIS, and RIS-user $k$, respectively, which are modelled as random matrices; $\beta_n=1$ is for all RIS elements; \eqref{eq-Pd} restricts the transmit power being not larger than $P_T$. 

This problem is an NP-hard non-convex mixed integer problem. Furthermore, fitness landscape analysis in \cite{yan2022fitness} revealed that the problem has a severe unstructured and rugged landscape, especially in cases with large-scale RIS elements. The common solver to the problem is using water-filling \cite{yu2004iterative} to obtain BS beamforming and estimating each RIS element's phase shift separately. The decoupled estimation has been demonstrated to be ineligible \cite{yan2022fitness}. Metaheuristics’ global search ability is potential to handle the unstructured, rugged, and highly coupled problem.

\subsubsection{Applying AutoOpt} We apply AutoOpt to design solvers to estimate RIS phase shifts. Problem instances with the number of RIS elements varies from 120 to 400 are considered. We randomly select five instances and target them during algorithm design; the other five are for comparing the designed algorithm with baselines. Other settings are the same as the setup in subsection \ref{sec_setup}.

Pseudocode of the designed algorithm (Alg$_{\rm beamform}$) is given in Algorithm \ref{alg_beamform}. Interestingly, the niching mechanism \texttt{choose\_nich} is preferred, which restricts the following uniform crossover (with crossover rate of $0.1229$) to be implemented within a neighborhood area. The one-dimensional reset operator further exploits the neighborhood area. Finally, the round-robin selection \texttt{update\_round\_robin} maintains diversity by keeping some inferior solutions. All these designs imply that maintaining solution diversity contributes to finding the global optima over the unstructured and rugged landscape.  
\begin{algorithm}[t]
\caption{Alg$_{\rm beamform}$}
\label{alg_beamform}
\footnotesize
\KwIn{initial solutions $S$}
\While{stopping criterion not met}{		
$S=\texttt{choose\_nich}(S)$ \\			
$S_{new}=\texttt{cross\_point\_uniform}(0.12,S)$ \\			
$S_{new}=\texttt{search\_reset\_one}(S_{new})$ \\
$S=\texttt{update\_round\_robin}(S,S_{new})$ \\		
}
\KwOut{optimal phase shifts from $S$}
\end{algorithm}

To investigate Alg$_{\rm beamform}$'s efficiency in the passive beamforming, we compare it with random beamforming, the representative sequential beamforming \cite{di2020hybrid}\footnote{Sequential beamforming refers to exhaustively enumerating the phase shift of each element one-by-one while keeping the remaining phase shifts unchanged.}, and three classic metaheuristic solvers, i.e., GA, ILS, and SA. The algorithms are compared on the five test instances. Results are given in Table \ref{tab_beamform}, in which the performance is measured by final solutions' fitness, i.e., reciprocal of the quality of service of all users. From Table \ref{tab_beamform}, sequential beamforming performs the worst in most cases, demonstrating the ineligibility of decoupling RIS elements. Alg$_{\rm beamform}$ outperforms other metaheuristic solvers, especially in larger-scale RIS cases (with more rugged landscapes). This performance benefits from its outstanding ability of diversity maintenance. All the above demonstrates the efficiency of AutoOpt's automated design techniques on the problem.
\begin{table*}[t]
\centering
\caption{Average and standard deviation of performance on beamforming. Best results are in bold.}
\label{tab_beamform}
\begin{tabular}{lccccc}
\toprule
\multirow{2}{*}{Algorithm} & \multicolumn{5}{c}{Number of RIS elements in the problem instances}\\
                       & 120 & 160 & 280 & 320 & 400\\
\midrule
Alg$_{\rm beamform}$ & \textbf{0.0332$\pm$5.05E-04} & \textbf{0.0312$\pm$4.84E-04} & \textbf{0.0281$\pm$1.57E-04} & \textbf{0.0272$\pm$6.76E-04} & \textbf{0.0260$\pm$1.11E-04} \\
Random                 & 0.0442$\pm$7.94E-04 & 0.0425$\pm$6.56E-04 & 0.0402$\pm$8.30E-04 & 0.0390$\pm$6.67E-04 & 0.0375$\pm$1.88E-04 \\
Sequential             & 0.0382$\pm$6.19E-04 & 0.0387$\pm$6.75E-04 & 0.0374$\pm$4.17E-04 & 0.0369$\pm$4.38E-04 & 0.0354$\pm$8.27E-04 \\
GA                     & 0.0369$\pm$3.30E-04 & 0.0356$\pm$1.00E-04 & 0.0337$\pm$4.26E-04 & 0.0333$\pm$1.04E-04 & 0.0322$\pm$6.96E-04 \\
ILS                    & 0.0333$\pm$3.74E-04 & 0.0314$\pm$2.49E-04 & 0.0285$\pm$1.19E-04 & 0.0279$\pm$1.82E-04 & 0.0278$\pm$1.15E-04 \\
SA                     & 0.0398$\pm$5.59E-04 & 0.0388$\pm$7.75E-04 & 0.0369$\pm$3.27E-04 & 0.0360$\pm$4.18E-04 & 0.0355$\pm$9.50E-04 \\
\bottomrule
\end{tabular}
\end{table*}

\subsection{Summary}
In this section, we have investigated AutoOpt's performance on numerical benchmark functions and real problems. These functions and problems are characterized by different landscapes, e.g., continuous, discrete, uni-modal, multi-modal, and non-separable. AutoOpt has designed eligible algorithms for these problems, which demonstrates AutoOpt's overall efficiency in different algorithm design scenarios. Furthermore, as can be seen from Algorithms \ref{alg_f1}, \ref{alg_f2}, \ref{alg_f9}, \ref{alg_restoration}, and \ref{alg_beamform}, the algorithms designed by AutoOpt are in various structures. These results confirm that suitable solvers to different problems may be disparate, such that designing algorithms within a fixed structure is insufficient. In this regard, it is necessary to have the general AutoOpt framework to discover various algorithm structures to fit different problem-solving.      

\section{Conclusions}\label{sec_conclusion}
In this paper, we have proposed the AutoOpt framework for automatically designing metaheuristic optimization algorithms with various structures. AutoOpt contains a general algorithm prototype, an acyclic graph algorithm representation, and graph representation embedding, which enables a high-performance algorithm design by fully discovering potentials, novelties, and diversity across the metaheuristic family. AutoOpt can be implemented with plenty of design methods and performance evaluation strategies to fit different algorithm design scenarios. Empirical studies on numerical functions and real problems have demonstrated AutoOpt's validity in discovering efficient algorithms with different structures. 

Future works include a further experimental analysis of the relationship between algorithm structures and problem features, a principled design method that maintains diversity in terms of algorithm structures and behaviors, and a computing architecture that utilizes distributed computing resources to speed up the algorithm design. Moreover, the extensive performance evaluations required make automated design computationally intensive. While the present work has contributed to the algorithm prototype and representation, it is crucial to explore performance evaluation strategies to either reduce the number of performance evaluations or estimating the performance without a full evaluation, thereby enhancing the computational efficiency of the automated design.

\bibliographystyle{IEEEtran}
\bibliography{References}

\end{document}